\documentclass[12pt,leqno,tbtags]{amsart}

\makeatletter

\usepackage{endnotes}
\let\footnote=\endnote
\def\notesname{Endnotes}

\def\enoteheading{\section*{\notesname
  \@mkboth{\uppercase{\notesname}}{\uppercase{\notesname}}}%
     \leavevmode\par}
\def\enoteformat{\rightskip\z@ \leftskip\z@ \parindent\z@
     \leavevmode$^{\@theenmark}$}

\usepackage{txfonts}

\newbox\barb@x
\newbox\barb@y
\newcommand{\leftbar}[2]{%
    \setbox\barb@x\hbox{$#1#2$}%
    \setbox\barb@y\hbox{$#1\mkern -2.735mu|$}%
    \mkern 1.585mu
    \kern -\wd\barb@y
    \vrule height\ht\barb@x depth\dp\barb@x width\wd\barb@y
    \setbox\barb@y\box\voidb@x
    \mkern -1.585mu
    \box\barb@x}
\newcommand{\rightbar}[2]{%
    \setbox\barb@x\hbox{$#1#2$}%
    \setbox\barb@y\hbox{$#1\mkern -2.735mu|$}%
    \copy\barb@x
    \mkern -1.585mu
    \vrule height\ht\barb@x depth\dp\barb@x width\wd\barb@y
    \kern -\wd\barb@y
    \setbox\barb@x\box\voidb@x
    \setbox\barb@y\box\voidb@x
    \mkern 1.585mu}

\calclayout

\def\@settitle{%
  \begingroup\centering
  \large\bfseries
  \@title
  \par\endgroup
}

\def\@setauthors{\normalfont\normalsize
  \begingroup\centering
  \vskip\linespacing
  \andify\authors
  \def\\{\protect\linebreak}%
  \authors
  \def\author##1{}%
  \def\address##1##2{\par\textit{%
    \@ifnotempty{##1}{\ignorespaces##1\unskip}{\ignorespaces##2\unskip}}}%
  \def\curraddr##1##2{}%
  \def\email##1##2{}%
  \def\urladdr##1##2{}%
  \addresses
  \par\endgroup
}

\renewenvironment{abstract}{%
  \ifx\maketitle\relax
    \ClassWarning{\@classname}{Abstract should precede
      \protect\maketitle\space in AMS documentclasses; reported}%
  \fi
  \global\setbox\abstractbox=\vtop \bgroup
    \normalfont\normalsize
    \list{}{\labelwidth\z@
      \leftmargin0pt \rightmargin\leftmargin
      \listparindent\normalparindent \itemindent\z@
      \parsep\z@ \@plus\p@
      
    }%
    \item[]%
}{%
  \endlist\egroup
  \ifx\@setabstract\relax \@setabstracta \fi
}

\def\@setabstracta{%
  \ifvoid\abstractbox
  \else
    \skip@2\linespacing \advance\skip@-\lastskip
    \advance\skip@-\baselineskip \vskip\skip@
    \box\abstractbox
    \prevdepth\z@ 
  \fi
}

\def\@setthanks{\def\thanks##1{\par##1\@addpunct.}\@thankses}

\def\@maketitle{%
  \normalfont\normalsize
  \let\@makefnmark\relax  \let\@thefnmark\relax
  \ifx\@empty\thankses\else \def\@theenmark{*} \global\let\@thankses\thankses
    \@endnotetext{\def\par{\let\par\@par}\@setthanks}\fi
  \@mkboth{\@nx\shortauthors}{\@nx\shorttitle}%
  \@settitle
  \ifx\@empty\authors \else \@setauthors \fi
  \ifx\@empty\@dedicatory
  \else
    \baselineskip18\p@
    \vtop{\centering{\footnotesize\itshape\@dedicatory\@@par}%
      \global\dimen@i\prevdepth}\prevdepth\dimen@i
  \fi
  \@setabstract
  \normalsize
  \if@titlepage
    \newpage
  \else
    \vskip\z@\relax
  \fi
} 

\let\enddoc@text\relax

\def\@sect#1#2#3#4#5#6[#7]#8{%
  \edef\@toclevel{\ifnum#2=\@m 0\else\number#2\fi}%
  \ifnum #2>\c@secnumdepth \let\@secnumber\@empty
  \else \@xp\let\@xp\@secnumber\csname the#1\endcsname\fi
  \@tempskipa #5\relax
  \ifnum #2>\c@secnumdepth
    \let\@svsec\@empty
  \else
    \refstepcounter{#1}%
    \edef\@secnumpunct{%
      \ifdim\@tempskipa>\z@ 
        \@ifnotempty{#8}{.\space\space}%
      \else .\space\space
      \fi
    }%
    \protected@edef\@svsec{%
      \ifnum#2<\@m
        \@ifundefined{#1name}{}{%
          \ignorespaces\csname #1name\endcsname\space
        }%
      \fi
      \@seccntformat{#1}%
    }%
  \fi
  \ifdim \@tempskipa>\z@ 
    \begingroup #6\relax
    \@hangfrom{\hskip #3\relax\@svsec}{\interlinepenalty\@M #8\par}%
    \endgroup
    \ifnum#2>\@m \else \@tocwrite{#1}{#8}\fi
  \else
  \def\@svsechd{#6\hskip #3\@svsec
    \@ifnotempty{#8}{\ignorespaces#8\unskip
       \@addpunct.}%
    \ifnum#2>\@m \else \@tocwrite{#1}{#8}\fi
  }%
  \fi
  \global\@nobreaktrue
  \@xsect{#5}}

\def\section{\def\@secnumfont{\bfseries}%
 \@startsection{section}{1}%
  \z@{-2\linespacing}{\linespacing}%
  {\normalfont\bfseries}}
\def\subsection{\def\@secnumfont{\mdseries}%
 \@startsection{subsection}{2}%
  \z@{-\linespacing}{\linespacing}%
  {\normalfont\itshape}}
\def\subsubsection{\def\@secnumfont{\mdseries}%
 \@startsection{subsubsection}{3}%
  \z@{-\linespacing}{\z@}%
  {\normalfont}}

\def\@adjustvertspacing{%
  \bigskipamount2\baselineskip
  \medskipamount\baselineskip
  \smallskipamount\z@
  \abovedisplayskip\baselineskip
  \belowdisplayskip\baselineskip
  \abovedisplayshortskip\baselineskip
  \belowdisplayshortskip\baselineskip
  \jot\z@ \relax
}
\listisep\z@
\captionindent\z@

\makeatother

\usepackage{url}
\usepackage{pstricks}
\usepackage{tree-dvips}
\usepackage{stmaryrd}

\usepackage[longnamesfirst,sort&compress]{natbib}
\newcommand{\citearound}[2]{{\let\realspace\ \def\ {\let\ \realspace#1}{#2}}}
\newcommand{\citegenitive}{\citearound{'s }}
\newcommand{\citets}[1]{\citegenitive{\citet{#1}}}

\usepackage{linguex}
\renewcommand{\refdash}{}
\makeatletter
\let\c@ExNo\c@equation
\let\theExNo\theequation
\renewcommand{\theSubExNo}{\hbox{\if@noftnote\arabic{ExNo}\else
   \roman{FnExNo}\fi\refdash\alph{SubExNo}}}
\renewcommand{\theSubSubExNo}{%
   \hbox{\if@noftnote\arabic{ExNo}\else\roman{FnExNo}\fi
          \refdash\alph{SubExNo}\refdash\if@noftnote\roman{SubSubExNo}%
                          \else\arabic{SubSubExNo}\fi}}
\makeatother

\usepackage{qtree}
\qtreecenterfalse
\newcommand{\roof}[2]{\leaf{\qroof{#1}.{#2} }}
\newcommand{\qobitreecenter}[1][]
    {\vcenter{\hbox{\raise-3.5pt\vbox{\hbox{\qobitree}\vskip0pt}#1}}}
\newcommand{\branchfakewidth}[2]
    {\branch{#1}{#2}\faketreewidth{#2}}

\makeatletter
\newcommand{\displayleft}[1]{\strut@#1\span\omit\column@plus\tabskip\alignsep@\hfill\\}
\newcommand{\displayright}[1]{\strut@#1\span\omit\column@plus\tabskip\alignsep@\\}
\makeatother

\newcommand{\sectionsign}{Section~}
\newcommand{\multisectionsign}{Sections~}
\newcommand{\sref}[1]{\sectionsign\ref{#1}}
\makeatletter
\newcommand{\srefrange}[3][.]{%
  \expandafter\ifx\csname r@#2\endcsname\relax
    \sref{#2}\nobreakdash--\ref{#3}%
  \else
  \expandafter\ifx\csname r@#3\endcsname\relax
    \sref{#2}\nobreakdash--\ref{#3}%
  \else
    \def\beforedot##1#1##2\end@fref{##1}%
    \def\afterdot ##1#1##2\end@fref{##2}%
    \def\finalize ##1#1\end@fref   {##1}%
    \edef\ref@ne{\ref{#2}#1}%
    \edef\reftw@{\ref{#3}#1}%
    \multisectionsign
    \def\next{%
      \expandafter\edef\expandafter\pre@ne
        \expandafter{\expandafter\beforedot\ref@ne\end@fref}%
      \expandafter\edef\expandafter\pretw@
        \expandafter{\expandafter\beforedot\reftw@\end@fref}%
      \ifx\pre@ne\pretw@
        \pre@ne#1%
        \expandafter\edef\expandafter\ref@ne
          \expandafter{\expandafter\afterdot\ref@ne\end@fref}%
        \expandafter\edef\expandafter\reftw@
          \expandafter{\expandafter\afterdot\reftw@\end@fref}%
        \ifx\@empty\ref@ne
          \ifx\@empty\reftw@
            \let\next\relax
          \else
            \def\next{\errmessage{The reference #2 has fewer #1-components than the reference #3.}}%
          \fi
        \else
          \ifx\@empty\reftw@
            \def\next{\errmessage{The reference #3 has fewer #1-components than the reference #2.}}%
          \fi
        \fi
      \else
        \expandafter\finalize\ref@ne\end@fref
        \nobreakdash--%
        \expandafter\finalize\reftw@\end@fref
        \let\next\relax
      \fi
      \next
    }%
    \next
  \fi\fi}
\makeatother

\makeatletter
\newenvironment{stacked}{%
  \begingroup
  \jot\z@
  
  \begin{aligned}
}{%
  \end{aligned}
  \endgroup
}
\makeatother

\newcommand{\denotation}[1]{\text{\scshape\lowercase{#1}}}
\newcommand{\Alice       }{\denotation{Alice}}
\newcommand{\Bob         }{\denotation{Bob}}
\newcommand{\Carol       }{\denotation{Carol}}
\newcommand{\Smoke       }{\denotation{smoke}}
\newcommand{\Love        }{\denotation{love}}
\newcommand{\Animate     }{\denotation{animate}}

\newcommand{\We          }{\denotation{we}}
\newcommand{\Buy         }{\denotation{buy}}
\newcommand{\Remember    }{\denotation{remember}}
\newcommand{\For         }{\denotation{for}}
\newcommand{\You         }{\denotation{you}}
\newcommand{\Think       }{\denotation{think}}

\newcommand{\eqish}[1]{\mathrel{\makebox[0pt]{\hphantom{$=$}\hss$#1$\hss}\hphantom{=}}}
\newcommand{\eqishcolon}{\eqish{:}}

\newcommand{\phrase}[1]{\emph{\frenchspacing#1}}
\newcommand{\denote}[1]{\ensuremath{\llbracket}#1\ensuremath{\rrbracket}}
\newcommand{\emp}[1]{\textsc{#1}}

\newcommand{\limplies}{\mathbin\Rightarrow}
\newcommand{\fun}[1]{\mathopen{\lambda\mathord{#1}.\,}}
\newcommand{\presuppose}[1]{\mathopen{[\mathord{#1}]\,}}
\newcommand{\Forall}[1]{\mathopen{\forall\mathord{#1}.\,}}
\newcommand{\Exists}[1]{\mathopen{\exists\mathord{#1}.\,}}

\newcommand{\pref}[1]{\ensuremath{\text{\ref{#1}}'}}

\newcommand{\toF}{\mathbin\rightarrow}
\newcommand{\toC}{\mathbin\rightharpoonup}
\newcommand{\toQ}{\mathbin\leadsto}

\newcommand{\Wh}{\protect\phrase{Wh}\protect\nobreakdash}
\newcommand{\wh}{\protect\phrase{wh}\protect\nobreakdash}

\newcommand{\cc}[2]{%
    \mathchoice{\ccDisplay{#1}{#2}}%
               {\ccInline{#1}{#2}}%
               {\ccInline{#1}{#2}}%
               {\ccInline{#1}{#2}}}
\newcommand{\ccDisplay}[2]{%
    \protect\bmatrix#2\\#1\endbmatrix}
\newcommand{\ccInline}[2]{%
    \bigl[\!\protect\smallmatrix#2\\#1\protect\endsmallmatrix\!\bigr]}

\newcommand{\branchaux}[3]{#1{\smash[b]{%
    \begin{tabular}[t]{@{}c@{}}%
        \begin{tabular}[b]{@{}c@{}}%
            $#3$%
        \end{tabular}%
        \\%
        #2%
    \end{tabular}}}}

\newcommand{\unarybranch}[1][]{\branchaux{\branch{1}}{\rlap{ $\scriptstyle#1$}}}
\newcommand{\binarybranch}[1][]{\branchaux{\branch{2}}{$\scriptstyle#1$}}
\newcommand{\unarybranchfakewidth}[1][]{\branchaux{\branchfakewidth{1}}{\rlap{ $\scriptstyle#1$}}}
\newcommand{\binarybranchfakewidth}[1][]{\branchaux{\branchfakewidth{2}}{$\scriptstyle#1$}}

\begin{document}

\title{A Continuation Semantics of Interrogatives\linebreak[1]
       That Accounts for Baker's Ambiguity}
\author{Chung-chieh Shan}
\address{Harvard University}
\email{ccshan@post.harvard.edu}
\urladdr{http://www.digitas.harvard.edu/\textasciitilde ken}
\date{\today}
\thanks{%
Thanks to Stuart Shieber, Chris Barker, Danny Fox,
Pauline Jacobson, Norman Ramsey, Dylan Thurston, MIT 24.979 Spring 2001
(Kai von Fintel and Irene Heim), the Harvard AI Research Group, the Center
for the Study of Language and Information at Stanford University, and the
referees at SALT~12.  This work is supported by National Science Foundation
Grant IRI-9712068.%
}

\begin{abstract}
\Wh-phrases in English can appear both raised and in-situ.  However, only
in\nobreakdash-situ \wh-phrases can take semantic scope beyond the immediately
enclosing clause.  I present a denotational semantics of interrogatives
that naturally accounts for these two properties.  It neither invokes
movement or economy, nor posits lexical ambiguity between raised and
in\nobreakdash-situ occurrences of the same \wh-phrase.  My analysis is based on the
concept of \emp{continuations}.  It uses a novel type system for
higher-order continuations to handle wide-scope \wh-phrases while remaining
strictly compositional.  This treatment sheds light on the
combinatorics of interrogatives as well as other kinds of so-called
\=A-movement.
\end{abstract}
\maketitle

\thispagestyle{empty}
\pagestyle{empty}

\section{Introduction}
\label{s:introduction}

\citet{baker-indirect} discusses multiple-\wh\ questions such
as those in~\eqref{e:baker-indirect}.
\ex. \label{e:baker-indirect}
     \a. \label{e:baker-indirect-1}
         Who remembers where we bought what?
     \b. \label{e:baker-indirect-2}
         Who do you think remembers what we bought for whom?\par
Each question in~\eqref{e:baker-indirect} contains three \wh-phrases and is
ambiguous between two readings with different notions of what constitutes
an appropriate answer.
\ex. Who remembers where we bought what?
     \a. \label{e:wide-scope-answer-1}
         Alice remembers where we bought the vase.
     \b. \label{e:narrow-scope-answer-1}
         Alice remembers where we bought what.\par
\ex. Who do you think remembers what we bought for whom?
     \a. \label{e:wide-scope-answer-2}
         I think Alice remembers what we bought for Bob.
     \b. \label{e:narrow-scope-answer-2}
         I think Alice remembers what we bought for whom.\par
Intuitively, both cases of ambiguity are because the final
\wh-phrase---\phrase{what} in~\eqref{e:baker-indirect-1} and \phrase{whom}
in~\eqref{e:baker-indirect-2}---can take either wide
scope~(\ref{e:wide-scope-answer-1}, \ref{e:wide-scope-answer-2}) or narrow
scope~(\ref{e:narrow-scope-answer-1}, \ref{e:narrow-scope-answer-2}).%
  \footnote{I disregard here the distinction between questions
  that allow or expect \emp{pair-list} answers (\phrase{Alice
  remembers where we bought the vase, and Bob remembers where we bought the
  table}) and questions that require or expect non-pair-list answers.}

In this paper, I focus on two properties of interrogatives.
\begin{itemize}
\item \label{i:puzzle-1} \Wh-phrases appear both raised and in-situ.  For
      example, in~\eqref{e:baker-indirect-2}, \phrase{who} and \phrase{what}
      appear raised while \phrase{whom} appears in-situ.
\item \label{i:puzzle-2} Raised \wh-phrases must take semantic scope exactly
      over the clause they are raised to overtly.  For example,
      in~\eqref{e:baker-indirect-2}, \phrase{who} must take wide scope, and
      \phrase{what} must take narrow scope.  Only \phrase{whom} has
      ambiguous scope; accordingly, the question has only $2$ readings, not
      $4$ or $8$.%
  \footnote{To be clear, a raised \wh-phrase can often take
  scope beyond the clause immediately enclosing its corresponding gap.
  The generalization I state here is that a raised \wh-phrase cannot take
  scope beyond the clause in front of which it is pronounced.}
\end{itemize}
I present a strictly compositional semantics of interrogatives in
English that accounts for these properties.  Specifically, in my analysis,
\begin{itemize}
\item there is no covert movement or \wh-raising between surface syntax and
      denotational semantics (contra \citets{epstein-derivational} economy
      account), yet a single denotation suffices for both raised and
      in-situ appearances of each \wh-phrase.  Moreover,
\item as a natural consequence of the denotation of \wh-phrases and the
      rules of the grammar, only in-situ \wh-phrases can take scope
      ambiguously.
\end{itemize}
I describe my system below as one where, roughly speaking, interrogative
clauses denote functions from answers to propositions (an old idea).
However, such
denotations are not crucial for my purposes---the essential ideas in my
analysis carry over easily to a system where interrogative clauses denote
say sets of propositions instead.  Hence this paper bears not so much
on what interrogatives denote, but how.

My analysis builds upon Barker's (\citeyear{barker-continuations},
\citeyear{barker-higher-order}) use of \emp{continuations} to characterize
quantification in natural language.  In~\sref{s:from-ptq}, I introduce
continuation semantics as a two-step generalization of
\citets{montague-proper} treatment of quantification.  The system I present
generalizes Barker's semantics in several aspects, which I point out below
as we encounter.  In~\sref{s:manipulating-answer-types}, I specify
denotations for interrogative elements and account for the properties
above.  In doing so, I am not concerned with the semantics of verbs such as
\phrase{know} that take interrogative complements, but rather with how to
derive denotations for interrogative clauses themselves.  Finally,
in~\sref{s:discussion}, I conclude with some speculations on further
applications of my treatment, for example to explain superiority effects.

This paper loosely follows the framework of Combinatory Categorial Grammar
(\citealp{steedman-combinatory}, \citeyear{steedman-surface}).  However,
I expect the central insights to be easy to adapt to other frameworks of
compositional semantics.

\section{From PTQ to Continuations in Two Steps}
\label{s:from-ptq}

Continuations are a well-known and widely applied idea in computer science.
Many analogies have been drawn to explain the concept; for example, in
programming language semantics, it is often said that ``the continuation
represents an entire (default) future for the computation''
{\citep{kelsey-r5rs}}.

From the perspective of natural language semantics, a continuation can be
thought of as ``a semantic value with a hole''.  To illustrate,
consider the sentence
\ex.\label{e:alice-loves-bob}Alice loves Bob.\par
In Montague grammar, the meaning of~\eqref{e:alice-loves-bob} is
computed compositionally from denotations assigned to
\phrase{Alice}, \phrase{loves}, and \phrase{Bob}.  Starting
with the values
\begin{equation}
    \Alice: e,\qquad
    \Love : e\toF e\toF t,\qquad
    \Bob  : e,
\end{equation}
where $e$ is the base type of individuals and $t$ is the base type of
propositions, we recursively combine semantic values by function
application (Figure~\ref{f:direct-application})
\begin{figure}[tp]
\centerline{$
    \leaf{$f : \alpha\toF\beta\qquad$}
    \leaf{$\qquad x : \alpha$}
    \branch{2}{$f(x) : \beta$}
    \qobitree
$}
\makeatletter
\caption{Function application%
    \hbox{\@textsuperscript{\normalfont\ref{f:type-variables}}}}
\makeatother
\label{f:direct-application}
\end{figure}
to obtain the top-level denotation $\Love(\Bob)(\Alice)$, of type~$t$
(Figure~\ref{f:alice-loves-bob}).%
    \footnote{\label{f:type-variables}%
    Throughout this paper, I use the Greek letters $\alpha$, $\beta$,
    $\gamma$, and~$\delta$ to represent type variables, in other words
    variables that can be instantiated with any type.  Thus what this paper
    calls types are known as \emp{type-schemes} in the Hindley-Milner type
    system \citep{hindley-combinators}.  Also, by convention, all binary
    type constructors associate to the right: The type $e\toF
    e\toF t$, unparenthesized, means $e\toF(e\toF t)$.}%
\begin{figure}[tp]
\centerline{$
    \leaf{$\Alice : e$}
    \leaf{$\Love : e\toF e\toF t$}
    \leaf{$\Bob : e$}
    \branch{2}{$\Love(\Bob) :e\toF t$}
    \branch{2}{$\Love(\Bob)(\Alice) :t$}
    \qobitree
$}
\caption{\protect\phrase{Alice loves Bob}}
\label{f:alice-loves-bob}
\end{figure}
Borrowing computer science terminology, I say that the
\emp{evaluation context} of the constituent \phrase{Bob}
in the sentence~\eqref{e:alice-loves-bob} is the ``sentence with a hole''
\ex.\label{e:alice-loves-hole}
    Alice loves~\_.\par
Semantically, this evaluation context is essentially a map
\begin{equation}
\label{e:alice-loves-cont}
    c = \fun{x} \Love(x)(\Alice)
\end{equation}
from individuals to propositions---in particular, the map~$c$ sends the
individual $\Bob$ to the proposition $\Love(\Bob)(\Alice)$.  The map~$c$ is
called the \emp{continuation} of \phrase{Bob}
in~\eqref{e:alice-loves-bob}.  I assign it the type
\begin{equation}
    e\toC t,
\end{equation}
where $\toC$ is a binary type constructor.

The \emp{value type} of a continuation is its domain, and the \emp{answer
type} of a continuation is its codomain.  For example, the continuation~$c$
has value type~$e$ and answer type~$t$.%
  \footnote{The concept of answer types in continuation semantics is
  separate from the concept of appropriate answerhood in
  interrogatives.}
I distinguish between the continuation type $e\toC t$ and the function type
$e\toF t$, even though they may be interpreted the same way
model-theoretically (standardly, as sets of functions).
In fact, I use the same notation to construct and apply functions (namely
$\lambda$ and parentheses) as for continuations.  The purpose of
distinguishing between continuations and functions is to maintain mental
hygiene and rule out undesirable semantic combination
(see \sref{s:raised-wh-phrases} below).

With the continuation~$c$ in~\eqref{e:alice-loves-cont} in hand, we can
apply it to $\Bob$ to recover the proposition that Alice loves Bob, or
apply it to $\Carol$ to generate the proposition that Alice loves Carol.
We can play ``what-if'' with the hole in~\eqref{e:alice-loves-hole},
plugging in different individuals to see what proposition the top-level
answer would come out to be.  In particular, we can compute
\begin{equation}
\label{e:everyone-intuition}
    \Forall{x} c(x)
\end{equation}
to generate the proposition that Alice loves every individual.  This
intuition is why \citets{montague-proper} ``Proper Treatment of
Quantification'' (PTQ) assigns essentially the denotation
\begin{equation}
\label{e:everyone-denotation}
    \denote{\phrase{everyone}} = \fun{c} \Forall{x} c(x)
    : (e\toC t)\toF t
\end{equation}
to the quantificational NP \phrase{everyone}.%
    \footnote{So do \citets{hendriks-studied} and
    \citets{barker-continuations} later proposals.  For simplicity and
    because they are irrelevant, I omit restrictions on quantification
    (``$\Animate(x)\limplies\dotsb$'') in~\eqref{e:everyone-denotation} and
    below.}

Note that the type in~\eqref{e:everyone-denotation} is $(e\toC t)\toF t$
rather than the more familiar $(e\toF t)\toF t$.  This type documents our
intuition that the denotation of \phrase{everyone} is a function that maps
each proposition with an $e$\nobreakdash-hole (type $e\toC t$) to
a proposition with no hole (type~$t$).  In general, a semantic value whose
type is of the form $\alpha\toC\gamma$ can be thought of as ``an~$\gamma$
with a $\alpha$\nobreakdash-hole'', in other words a continuation
that---given a hole-filler of type~$\alpha$---promises to produce an answer
of type~$\gamma$.  To redeem this promise is to feed the continuation to
a function of type $(\alpha\toC\gamma)\toF\gamma'$ in return for a final
answer of type~$\gamma'$.  (In the case of \phrase{everyone}
in~\eqref{e:everyone-denotation}, the two answer types $\gamma$
and~$\gamma'$ are both~$t$, and the value type~$\alpha$ is~$e$.)

As one might expect from this discussion, types of the form
$(\alpha\toC\gamma)\toF\gamma'$ recur throughout this paper.  I write
$\alpha\cc{\gamma}{\gamma'}$
as shorthand for such a type.  For example, the denotation of
\phrase{everyone} given in~\eqref{e:everyone-denotation} can be
alternatively written
\begin{equation}
    \denote{\phrase{everyone}} = \fun{c} \Forall{x} c(x) : e\ccInline{t}{t}.
\end{equation}
I call $\alpha$ the \emp{value type}, $\gamma$ the \emp{incoming answer
type}, and $\gamma'$ the \emp{outgoing answer type}.

Continuation semantics can be understood as a generalization of PTQ, in
two steps:
\begin{itemize}
\item Lift not just the semantic type of NPs from~$e$ to $e\cc{t}{t}$,
      but also the semantic type of other phrases from say~$\alpha$ to
      $\alpha\cc{t}{t}$.
\item Lift each type~$\alpha$ to not just the type $\alpha\cc{t}{t}$, but
      any type of the form $\alpha\cc{\gamma}{\gamma'}$, where $\gamma$
      and~$\gamma'$ are types.
\end{itemize}
I detail these steps below.

\subsection{First Generalization: From $e\cc{t}{t}$ to $\alpha\cc{t}{t}$}
\label{s:generalization-1}

In PTQ, the semantic type of NPs is not~$e$ but $e\cc{t}{t}$.
For example, the NPs \phrase{Alice} and \phrase{Bob} denote not the
individuals $\Alice:e$ and $\Bob:e$, but rather the lifted values
\begin{subequations}
\begin{alignat}{2}
    \denote{\phrase{Alice}} &= \fun{c} c(\Alice) &&: e\ccInline{t}{t},\\
    \denote{\phrase{Bob}}   &= \fun{c} c(\Bob)   &&: e\ccInline{t}{t}.
\end{alignat}
\end{subequations}
In general, any value~$x:e$ can be lifted to the value $\fun{c}
c(x):e\cc{t}{t}$.  The lifted type~$e\cc{t}{t}$ is borne by all NPs,
from proper names like \phrase{Alice} and \phrase{Bob} to quantificational
NPs such as \phrase{everyone} and \phrase{someone}.
\begin{subequations}
\label{e:everyone-someone-denotation}
\begin{alignat}{2}
    \denote{\phrase{everyone}} &= \fun{c} \Forall{x} c(x) &&: e\ccInline{t}{t},\\
    \denote{\phrase{someone}}  &= \fun{c} \Exists{x} c(x) &&: e\ccInline{t}{t}.
\end{alignat}
\end{subequations}
Although proper names and quantificational NPs share the same lifted type,
the denotations of the latter do not result from lifting any value.

PTQ is appealing in part because it assigns the same lifted type to all
NPs, quantificational or not.  We can generalize this uniformity beyond
NPs.  For example, let us lift intransitive verbs from the type $e\toF
t$ to $(e\toF t)\cc{t}{t}$, and transitive verbs from the type
$e\toF e\toF t$ to $(e\toF e\toF t)\cc{t}{t}$.
\begin{subequations}
\begin{alignat}{2}
    \label{e:smoke-denotation}
    \denote{\phrase{smoke}} &= \fun{c} c(\Smoke) &&: (e\toF t)\ccInline{t}{t},\\
    \denote{\phrase{love}}  &= \fun{c} c(\Love)  &&: (e\toF e\toF t)\ccInline{t}{t}.
\end{alignat}
\end{subequations}
In general, any semantic value~$x$, say of type~$\alpha$, can be lifted to
the value $\fun{c} c(x)$, of type~$\alpha\cc{t}{t}$.  This lifting rule
is shown in Figure~\ref{f:lifting-t}.
\begin{figure}[tp]
\centerline{$
    \leaf{$x : \alpha$}
    \branch{1}{$\fun{c} c(x) : \alpha\cc{t}{t}$}
    \qobitree
$}
\caption{Lifting semantic values}
\label{f:lifting-t}
\end{figure}

To maintain the uniformity of types across the grammar, we want every VP to
take the same semantic type $(e\toF t)\cc{t}{t}$.  Furthermore, just
as the new denotation of \phrase{smoke} in~\eqref{e:smoke-denotation} is
its old denotation $\Smoke$ lifted, the new denotation of \phrase{love Bob}
should also be its old denotation, $\Love(\Bob)$, lifted.  What we now need
is a semantic rule that will combine a lifted function with a lifted
argument to form a lifted result.  For example, the rule should combine
$\denote{\phrase{love}} = \fun{c} c(\Love)$ with $\denote{\phrase{Bob}}
= \fun{c} c(\Bob)$ to form $\fun{c} c(\Love(\Bob))$, the denotation we
desire for \phrase{love Bob}.  This situation is depicted in
Figure~\ref{f:desired-fa}.
\begin{figure}[tp]
\centerline{$
    \leaf{$
        \denote{\phrase{love}} = \fun{c} c(\Love) : (e\toF e\toF t)\cc{t}{t}
    $}
    \leaf{$
        \denote{\phrase{Bob}} = \fun{c} c(\Bob) : e\cc{t}{t}
    $}
    \binarybranch[\raisebox{-2ex}{\framebox{\makebox[\totalheight]?}}]{
        \denote{\phrase{love Bob}}
        = \fun{c} c(\Love(\Bob))
        : (e\toF t)\cc{t}{t}}
    \qobitree
$}
\caption{The desired output of a lifted semantic rule}
\label{f:desired-fa}
\end{figure}

Consider now the following calculation.
\begin{equation}
\label{e:rewriting-1}
\begin{split}
    \fun{c} c(\Love(\Bob)) &= \fun{c} \bigl(\fun{c'}c'(\Love)\bigr) \bigl(\fun{f} c(f(\Bob))\bigr)
    \\                     &= \fun{c} \denote{\phrase{love}} \bigl(\fun{f} c(f(\Bob))\bigr)
    \\                     &= \fun{c} \denote{\phrase{love}} \bigl(\fun{f} \bigl(\fun{c'} c'(\Bob)\bigr) \bigl(\fun{x} c(f(x))\bigr) \bigr)
    \\                     &= \fun{c} \denote{\phrase{love}} \bigl(\fun{f} \denote{\phrase{Bob}} \bigl(\fun{x} c(f(x))\bigr) \bigr).
\end{split}
\end{equation}
In the first two lines, the atom~$\Love$ is replaced with a
variable~$f$, which gets its value from the lifted denotation of
\phrase{love}.  In the last two lines, the atom~$\Bob$ is similarly
replaced with~$x$, which gets its value from the
lifted denotation of \phrase{Bob}.  The end result is a way to write down
the lifted result of a function application in terms of the lifted function
and the lifted argument, without mentioning any unlifted atoms.  This
technique generalizes to a new semantic rule, \emp{lifted
function application}, shown in Figure~\ref{f:lifted-fa-1-t}.  It satisfies
the requirement in Figure~\ref{f:desired-fa}, as is easily checked.
\begin{figure}[tp]
\centerline{$
    \leaf{$\bar{f} : (\alpha\toF\beta)\cc{t}{t}\qquad$}
    \leaf{$\qquad\bar{x} : \alpha\cc{t}{t}$}
    \branch{2}{$
          \fun{c} \bar{f}\bigl(\fun{f}
                  \bar{x}\bigl(\fun{x}
                  c(f(x))\bigr)\bigr) : \beta\cc{t}{t}$}
    \qobitree
$}
\caption{Lifted function application
    (evaluating function then argument)}
\label{f:lifted-fa-1-t}
\end{figure}

As it turns out, there is another way to satisfy the requirement.
The calculation
in~\eqref{e:rewriting-1} above replaces the atom~$\Love$ first and the
atom~$\Bob$ second.  If instead we replace $\Bob$ first and $\Love$
second, we arrive at a different result.
\begin{equation}
\label{e:rewriting-2}
\begin{split}
    \fun{c} c(\Love(\Bob)) &= \fun{c} \bigl(\fun{c'}c'(\Bob)\bigr) \bigl(\fun{x} c(\Love(x))\bigr)
    \\                     &= \fun{c} \denote{\phrase{Bob}} \bigl(\fun{x} c(\Love(x))\bigr)
    \\                     &= \fun{c} \denote{\phrase{Bob}} \bigl(\fun{x} \bigl(\fun{c'} c'(\Love)\bigr) \bigl(\fun{f} c(f(x))\bigr) \bigr)
    \\                     &= \fun{c} \denote{\phrase{Bob}} \bigl(\fun{x} \denote{\phrase{love}} \bigl(\fun{f} c(f(x))\bigr) \bigr).
\end{split}
\end{equation}
This alternative calculation in turn gives rise to a different lifted
function application rule, that in Figure~\ref{f:lifted-fa-2-t}.
\begin{figure}[tp]
\centerline{$
    \leaf{$\bar{f} : (\alpha\toF\beta)\cc{t}{t}\qquad$}
    \leaf{$\qquad\bar{x} : \alpha\cc{t}{t}$}
    \branch{2}{$
          \fun{c} \bar{x}\bigl(\fun{x}
                  \bar{f}\bigl(\fun{f}
                  c(f(x))\bigr)\bigr)
        : \beta\cc{t}{t}$}
    \qobitree
$}
\caption{Lifted function application
    (evaluating argument then function)}
\label{f:lifted-fa-2-t}
\end{figure}

The two rules in Figures~\ref{f:lifted-fa-1-t}
and~\ref{f:lifted-fa-2-t} differ in \emp{evaluation order}.  Roughly
speaking, the evaluation order of a programming language is the order in
which ``computational side effects'' like input and output occur as
expressions are evaluated (in other words as code is executed).
Continuations are often used in programming language semantics to model
evaluation order, for instance in \citets{plotkin-call} seminal work.
Adopting this terminology, I say that our first rule evaluates the function
before the argument, and our second rule evaluates the argument before the
function.

\subsection{Quantification}

With the semantic rules and lexical denotations introduced so far, we can
derive the sentence \phrase{Alice loves everyone}.
Figure~\ref{f:alice-loves-everyone} shows one analysis, essentially that of
\citets{barker-continuations}.
\begin{figure}[tp]
\centerline{$
    \leaf{$\Alice:e$}
    \unarybranch[\wedge]{\fun{c} c(\Alice) :e\cc{t}{t}}
    \leaf{$\Love:e\toF e\toF t$}
    \unarybranch[\wedge]{\fun{c} c(\Love) :(e\toF e\toF t)\cc{t}{t}}
    \leaf{$\fun{c} \Forall{x} c(x):e\cc{t}{t}$}
    \binarybranchfakewidth[>]{\fun{c} \Forall{x} c(\Love(x)) :(e\toF t)\cc{t}{t}}
    \binarybranch[>]{\fun{c} \Forall{x} c(\Love(x)(\Alice)) : t\cc{t}{t}}
    \qobitree
$}
\caption{\protect\phrase{Alice loves everyone}}
\label{f:alice-loves-everyone}
\end{figure}

I indicate semantic rules used in derivations with the following notation.
\begin{itemize}
\item A unary branch decorated with~$\wedge$ invokes the lifting rule
      (Figure~\ref{f:lifting-t}).
\item A binary branch decorated with~$>$ invokes the one of the two lifted
      function application rules that evaluates the left
      daughter first, in other words either the
      function-then-argument rule (Figure~\ref{f:lifted-fa-1-t})
      or the mirror image of the argument-then-function rule
      (Figure~\ref{f:lifted-fa-2-t}).
\item A binary branch decorated with~$<$ invokes the other of the two
      lifted function application rules, which evaluates
      the right daughter first.
\end{itemize}
In this derivation, all evaluation orders give the same result, so both
binary branches in Figure~\ref{f:alice-loves-everyone} could have been
decorated with~$<$ instead of~$>$.  Evaluation order does not matter for
this sentence because it only contains one quantificational NP\@.  For
other sentences, such as \phrase{someone loves everyone}, different orders
of evaluation give differently scoped results \citep{barker-continuations}.

\subsection{Lowering}

The top-level denotation derived in Figure~\ref{f:alice-loves-everyone} has
type~$t\cc{t}{t}$, in other words $(t\toC t)\toF t$.  This type is not~$t$,
the type that clauses usually receive in Montague grammar.  This
discrepancy is expected, since every type~$\alpha$, say $\alpha=t$, has been
lifted to the type $\alpha\cc{t}{t}=t\cc{t}{t}$.  We can recover the usual,
type\nobreakdash-$t$ denotation of the sentence
by applying the lifted denotation---as a function---to the identity continuation.
That is, compute
\begin{equation}
\bigl( \fun{c} \Forall{x} c(\Love(x)(\Alice)) \bigr)
\bigl( \fun{x} x \bigr)
= \Forall{x} \Love(x)(\Alice) : t
\end{equation}
to recover the propositional meaning of \phrase{Alice loves everyone}.
Here the identity continuation $\fun{x} x : t\toC t$ corresponds to the
trivial evaluation context~``$\_$'': a sentence with a sentence hole, and
nothing else.

In general, from any value~$\bar{x}$ whose type is~$t\cc{t}{t}$ we can
recover the propositional value $\bar{x}(\fun{x}x)$, of type~$t$.
This gives us the \emp{lowering} rule in Figure~\ref{f:lowering-t}.
\begin{figure}[tp]
\centerline{$
    \leaf{$\bar{x} : t\cc{t}{t}$}
    \unarybranch[\vee]{\bar{x}(\fun{x}x) : t}
    \qobitree
$}
\caption{Lowering lifted propositions}
\label{f:lowering-t}
\end{figure}
Where this rule is used in derivations below, I decorate the unary branch
with~$\vee$.

\subsection{Second Generalization: From $\alpha\cc{t}{t}$ to $\alpha\cc{\gamma}{\gamma'}$}
\label{s:generalization-2}

The type~$t$ plays a special role in PTQ: It is the only answer type.
Because PTQ is after all a treatment of quantification, it is natural there
for the answer type to be fixed at~$t$ by lexical denotations like those of
\phrase{everyone} and \phrase{someone}
in~\eqref{e:everyone-someone-denotation}.  However, in the treatment of
extraction and interrogation that I present below, not all clauses have the
same type.  Gapped and interrogative clauses do not denote propositions;
they have types other than~$t$.  So our semantic rules must deal with
lifted values whose answer types are not~$t$.

\begin{figure}[tp]
\centerline{$
    \leaf{$x : \alpha$}
    \unarybranch[\wedge]{\fun{c} c(x) : \alpha\cc{\gamma}{\gamma}}
    \qobitree
$}
\caption{Lifting semantic values, revised}
\label{f:lifting}
\end{figure}

The semantic rules introduced so far (Figures~\ref{f:lifting-t},
\ref{f:lifted-fa-1-t}, \ref{f:lifted-fa-2-t}, and~\ref{f:lowering-t}) do not
mention any logical operator.  Thus the type~$t$ plays no essential role in
these rules, and can be replaced with a type variable~$\gamma$.  Start with
the lifting rule (Figure~\ref{f:lifting-t}): Any semantic value~$x$, say of
type~$\alpha$, can be lifted to the value~$\fun{c}c(x)$, of (polymorphic)
type~$\alpha\cc{\gamma}{\gamma}$.  This revised lifting rule is shown in
Figure~\ref{f:lifting}.

\begin{figure}[tp]
\centerline{$
    \leaf{$\bar{f} : (\alpha\toF\beta)\cc{\gamma_1}{\gamma_0}\qquad$}
    \leaf{$\qquad\bar{x} : \alpha\cc{\gamma_2}{\gamma_1}$}
    \binarybranch[>]{
          \fun{c} \bar{f}\bigl(\fun{f}
                  \bar{x}\bigl(\fun{x}
                  c(f(x))\bigr)\bigr) : \beta\cc{\gamma_2}{\gamma_0}}
    \qobitree
$}
\caption{Lifted function application
    (evaluating function then argument), revised}
\label{f:lifted-fa-1}
\end{figure}

\begin{figure}[tp]
\centerline{$
    \leaf{$\bar{f} : (\alpha\toF\beta)\cc{\gamma_2}{\gamma_1}\qquad$}
    \leaf{$\qquad\bar{x} : \alpha\cc{\gamma_1}{\gamma_0}$}
    \binarybranch[<]{
          \fun{c} \bar{x}\bigl(\fun{x}
                  \bar{f}\bigl(\fun{f}
                  c(f(x))\bigr)\bigr) : \beta\cc{\gamma_2}{\gamma_0}}
    \qobitree
$}
\caption{Lifted function application
    (evaluating argument then function), revised}
\label{f:lifted-fa-2}
\end{figure}

\begin{figure}[tp]
\centerline{$
    \leaf{$\bar{x} : \alpha\cc{\alpha}{\gamma}$}
    \unarybranch[\vee]{\bar{x}(\fun{x}x) : \gamma}
    \qobitree
$}
\caption{Lowering lifted values, revised}
\label{f:lowering}
\end{figure}

The other semantic rules can be similarly revised, by substituting~$\gamma$
for~$t$ throughout.  However, further generalization is possible.  We can
not only support answer types other than~$t$, but also allow multiple
answer types to occur in the same derivation.  In technical terms, consider
the $\lambda$\nobreakdash-terms in our semantic rules: How polymorphic can
their types be without risking a mismatch?  The most general types that can
be assigned are shown in Figures
\ref{f:lifted-fa-1}\nobreakdash--\ref{f:lowering}.  (Note that the two
versions of lifted function application now differ in their types.)

Let me summarize the semantics we have arrived at.  Alongside of function
application, we have added to Montague grammar four semantic rules:
lifting, two versions of lifted function application, and lowering.  These
four rules, shown in Figures \ref{f:lifting}\nobreakdash--\ref{f:lowering},
suffice below to analyze extraction and interrogation, except \wh-phrases
taking wide scope call for higher-order continuations
(\sref{s:higher-order}).

\section{Manipulating Answer Types}
\label{s:manipulating-answer-types}

In this section, I present my analysis of interrogatives using
continuations.  I first analyze extraction and raised \wh-phrases, then
turn to in-situ \wh-phrases and multiple-\wh\ interrogatives.

\subsection{Extraction}
\label{s:extraction}

\begin{figure}[tp]
\centerline{$
    \leaf{$\We:e$}
    \unarybranch[\wedge]{\fun{c} c(\We) : e\cc{e\toC t}{e\toC t}}
    \leaf{$\Buy:e\toF e\toF t$}
    \unarybranch[\wedge]{\fun{c} c(\Buy) :(e\toF e\toF t)\cc{e\toC t}{e\toC t}}
    \leaf{$\fun{c}c:(e\toC t)\toF (e\toC t)$}
    \binarybranchfakewidth[>]{\fun{c} \fun{x} c(\Buy(x)) : (e\toF t)\cc{t}{e\toC t}}
    \binarybranchfakewidth[>]{\fun{c} \fun{x} c(\Buy(x)(\We)) : t\cc{t}{e\toC t}}
    \unarybranch[\vee]{\fun{x} \Buy(x)(\We) : e\toC t}
    \qobitree
$}
\caption{\protect\phrase{We bought~\_}}
\label{f:we-bought-t}
\end{figure}

In order to analyze interrogatives with raised \wh-phrases, I first need a
theory of extraction.  Fortunately, continuation semantics provides for a
natural and compositional analysis of extraction.  One possible
implementation is to posit a phonologically null NP, notated~``\_'',
whose denotation is%
  \footnote{That I posit a phonologically null element
  is a matter of presentation and not critical to the approach to
  extraction sketched here.  It would work equally well for my purposes to
  introduce type-shift operations that effectively roll
  $\denote{\phrase{\_}}$ into binary rules.}
\begin{equation}
\label{e:trace-denotation}
    \denote{\phrase{\_}} = \fun{c} c : (e\toC\gamma)\toF(e\toC\gamma),
\end{equation}
that is, the identity function over individual-taking continuations.%
  \footnote{The idea that a variable or gap is in some sense an identity
  function over continuations appeared in the work of
  \citeauthor{danvy-functional} (\citeyear{danvy-functional}, \S3.4),
  and also has been mentioned to me by
  Barker.  The more general idea that a variable or gap is
  an identity function of some sort has an even longer history in computer
  science \citep{hindley-combinators} and linguistics
  \citep{jacobson-variable-free}.}
Figure~\ref{f:we-bought-t} shows how to
derive \phrase{we bought~\_}.  The final denotation has type $e\toC t$,
a continuation type.  This is typical of a clause with an unsaturated gap,
and is intuitive considering that continuations are supposed to model
evaluation contexts, in other words expressions with holes.

Figure~\ref{f:we-bought-t} reflects one important difference between the
continuation semantics here and Barker's system.  Barker takes
the answer type to be~$t$ everywhere; he does not generalize PTQ to
arbitrary answer types (as I do in \sref{s:generalization-2} above).
My present purposes require that we not fix a single answer type, for two
reasons.  First, not all clauses have the same type.  Although basic
declarative sentences such as \phrase{Alice loves Bob} have type~$t$,
I want sentences with gaps and interrogatives to have other semantic types.
Second, my semantics not only calls for a mixture of different answer
types, but in fact contains denotations that modify what can be
thought of as the ``current answer type''.

Manipulation of answer types is exemplified by the denotation
of~\phrase{\_} in~\eqref{e:trace-denotation} and the derivation of
\phrase{we bought~\_} in Figure~\ref{f:we-bought-t}.  Examining the
denotation of~\phrase{\_}, we see that it takes as input an
$e$\nobreakdash-taking
continuation whose answer type is $\gamma$, but returns a final answer of
type $e\toC\gamma$ instead.  Informally speaking, $\denote{\phrase{\_}}$
acts like an $e$ locally, but in addition prepends ``$e\>\toC$'' to the
current answer type; hence \phrase{we bought~\_} receives the semantic type
$e\toC t$ rather than $t$.  In general, a value of type
\begin{equation}
\label{e:manipulation}
    (\alpha\toC\gamma)\toF\gamma'
\end{equation}
acts like the value type~$\alpha$ locally, but in addition transforms the
incoming answer type~$\gamma$ to the outgoing answer type~$\gamma'$.  The
denotation $\denote{\phrase{\_}}$ is a special case where $\alpha = e$ and
$\gamma' = e\toC\gamma$.  Another special case is denotations lifted using
the lifting rule, for which $\gamma=\gamma'$ and manipulation of
answer types degenerates into propagation.%
  \footnote{This pseudo-operational description is merely an intuitive
  sketch.  The denotations in my semantics are computed purely in-situ
  according to local composition rules.}

The intuition that values like $\denote{\phrase{\_}}$ ``change the current
answer type'' is reflected my shorthand notation for continuation types,
introduced above in \sref{s:from-ptq}.  The type of $\denote{\phrase{\_}}$
can be written alternatively as~$e\cc{\gamma}{e\toC\gamma}$, so as to
emphasize that it acts locally like an $e$, but prepends ``$e\>\toC$'' to
the answer type.  Note also, in Figures~\ref{f:lifted-fa-1}
and~\ref{f:lifted-fa-2}, how lifted function application concatenates two
changes to the answer type---first from~$\gamma_2$ to~$\gamma_1$, and then
from~$\gamma_1$ to~$\gamma_0$---into a change from $\gamma_2$ to
$\gamma_0$.

\subsection{Raised \protect\Wh-phrases}
\label{s:raised-wh-phrases}

As alluded to in~\sref{s:introduction}, my interrogative denotations
are roughly functions mapping answers to propositions.  To
make this idea precise, I introduce yet another binary type constructor
$\toQ$, so as to form \emp{question types} such as $e\toQ t$.
As before, I distinguish between the question type $e\toQ t$, the
continuation type $e\toC t$, and the function type $e\toF t$, even though
they may have the same models and I overload the same notation to construct
and apply all three kinds of abstractions.

I now analyze the sentences
\ex. \a. \label{e:john-remembers-what-we-bought-t}
         Alice remembers what [we bought~\_].
     \b. What did we buy~\_?\par
I ignore the subject-auxiliary inversion triggered by direct (top-level)
interrogatives.

In~\sref{s:extraction} is derived a denotation of type $e\toC t$ for the
embedded clause \phrase{we bought~\_}.  Let us assume, as is commonly done,
that remembrance relates persons (type $e$) and questions (type $e\toQ t$
for now).  Then \phrase{remember} has type $(e\toQ t)\toF e\toF t$.

Having assigned meanings to every other word
in~\eqref{e:john-remembers-what-we-bought-t}, I need to specify
what \phrase{what} means.  Note that \phrase{we bought~\_} is of type $e\toC
t$, but \phrase{remember} requires the distinct type $e\toQ t$ for its
input.  Therefore, the denotation of \phrase{what} should convert
\phrase{we bought~\_} from $e\toC t$ to $e\toQ t$.  I make the simplest
assumption to that effect---that \phrase{what} has the semantic type
$(e\toC t) \toF (e\toQ t)$.
The semantic content of \phrase{what} should be essentially the identity
function, but express the requirement that the
input to $e\toQ t$ be somehow inanimate.  I am not
concerned with the nature of this requirement here,
so I simply notate it as a bracketed formula
$\presuppose{\neg\Animate(x)}\!$, as in%
\begin{subequations}
\label{e:wh-proto-denotations}
\begin{alignat}{2}
\label{e:what-proto-denotation}
    \denote{\phrase{what}}
          &= \fun{c} \fun{x} \presuppose{\neg\Animate(x)} c(x)
         &&: (e\toC t) \toF (e\toQ t),\\
\label{e:who-proto-denotation}
    \denote{\phrase{who}}
          &= \fun{c} \fun{x} \presuppose{\Animate(x)} c(x)
         &&: (e\toC t) \toF (e\toQ t).
\end{alignat}
\end{subequations}
The formula $\presuppose{\neg\Animate(x)} c(x)$
can be thought of as ``if $\neg\Animate(x)$ then $c(x)$, otherwise undefined''.

\begin{figure}[tp]
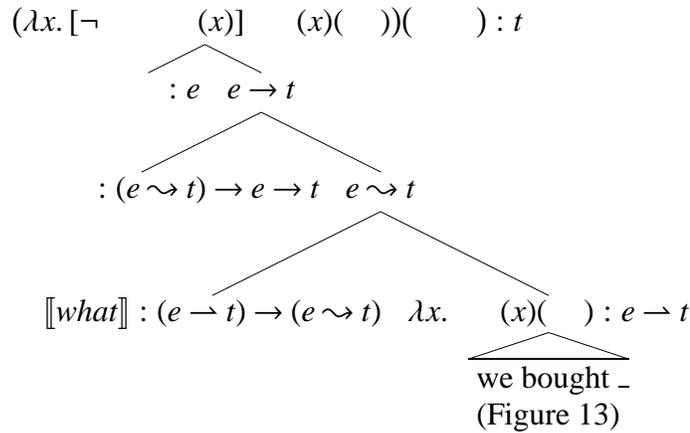

\centerline{$
    \leaf{$\Alice:e$}
    \leaf{$\Remember:(e\toQ t)\toF e\toF t$}
    \leaf{$\denote{\phrase{what}}:(e\toC t)\toF (e\toQ t)$}
    \roof{we bought~\_\\(Figure~\ref{f:we-bought-t})}{$
        \fun{x} \Buy(x)(\We) : e\toC t
      $}
    \binarybranchfakewidth{e\toQ t}
    \binarybranchfakewidth{e\toF t}
    \binarybranch{
        \Remember\bigl(\fun{x} \presuppose{\neg\Animate(x)} \Buy(x)(\We)\bigr)
                 \bigl(\Alice\bigr)
        : t
      }
    \qobitree
$}
\caption{\protect\phrase{Alice remembers what} [\protect\phrase{we bought~\_}]}
\label{f:john-remembers-what-we-bought-t}
\end{figure}

The definitions in~\eqref{e:wh-proto-denotations} makes no concrete use of the
base type~$t$.  Indeed, we can generalize them to%
\begin{subequations}
\label{e:wh-denotations}
\begin{alignat}{2}
\label{e:what-denotation}
    \denote{\phrase{what}}
          &= \fun{c} \fun{x} \presuppose{\neg\Animate(x)} c(x)
         &&: (e\toC \gamma) \toF (e\toQ \gamma),\\
\label{e:who-denotation}
    \denote{\phrase{who}}
          &= \fun{c} \fun{x} \presuppose{\Animate(x)} c(x)
         &&: (e\toC \gamma) \toF (e\toQ \gamma).
\end{alignat}
\end{subequations}
I use this generalization to analyze multiple-\wh\ clauses in
\sref{s:in-situ-wh-phrases} below.  Regardless, we can derive
\eqref{e:john-remembers-what-we-bought-t}.  One derivation is shown in
Figure~\ref{f:john-remembers-what-we-bought-t}.  Once the meaning of
\phrase{we bought~\_} is derived, the rest of the derivation consists
entirely of (unlifted) function application.

Why distinguish between function types
($\toF$), continuation types ($\toC$), and question types ($\toQ$)\@?
The distinction prevents the grammar from
overgenerating sentences like *\phrase{I remember Alice bought} or
*\phrase{I remember what what Alice bought}.  The types
enforce a one-to-one correspondence between gaps and \wh-phrases---more
precisely, interrogatives gaps and raised \wh-phrases.

\subsection{In-situ \protect\Wh-phrases}
\label{s:in-situ-wh-phrases}

The analyses above of extraction and raised \wh-phrases are both natural in
the spirit of continuation semantics.  The primary payoff from these
analyses is that little more needs to be said to treat interrogatives with
in-situ \wh-phrases and to account for the two properties I listed
in~\sref{s:introduction}.

\begin{figure}[tp]
\centerline{$
    \leaf{$
      \begin{stacked}
        \displayleft{\denote{\phrase{what}}:(e\toC e\toQ t)}
        \displayright{ \toF (e\toQ e\toQ t)}
      \end{stacked}$}
    \leaf{$\We:e$}
    \unarybranch[\wedge]{e\cc{e\toC e\toQ t}{e\toC e\toQ t}}
    \leaf{$\Buy:e\toF e\toF t$}
    \unarybranch[\wedge]{(e\toF e\toF t)\cc{e\toC e\toQ t}{e\toC e\toQ t}}
    \leaf{$\hspace{-2em}\fun{c}c$\\${}:e\cc{e\toQ t}{e\toC e\toQ t}$}
    \binarybranchfakewidth[>]{\fun{c} \fun{x} c(\Buy(x)) : (e\toF t)\cc{e\toQ t}{e\toC e\toQ t}}
    \leaf{$\For:e\toF(e\toF t)\toF(e\toF t)$}
    \unarybranchfakewidth[\wedge]{\hspace{-2em}\bigl(e\toF(e\toF t)$\\${}\toF(e\toF t)\bigr)\cc{e\toQ t}{e\toQ t}}
    \leaf{$\denote{\phrase{who}}:e\cc{t}{e\toQ t}$}
    \binarybranchfakewidth[>]{\qquad\bigl((e\toF t)\toF(e\toF t)\bigr)\cc{t}{e\toQ t}\quad}
    \binarybranchfakewidth[>]{(e\toF t)\cc{t}{e\toC e\toQ t}}
    \binarybranchfakewidth[>]{
        \rlap{$\fun{c} \fun{x} \fun{y} \presuppose{\Animate(y)}$}\qquad\qquad$\\$
        \rlap{$c(\For(y)(\Buy(x))(\We)) : t\cc{t}{e\toC e\toQ t}$}\qquad}
    \unarybranchfakewidth[\vee]{
        \rlap{$\fun{x} \fun{y} \presuppose{\Animate(y)}$}\qquad\qquad$\\$
        \rlap{$\For(y)(\Buy(x))(\We) :e\toC e\toQ t$}\qquad}
    \binarybranch{\rlap{$\fun{x} \presuppose{\neg\Animate(x)}
                         \fun{y} \presuppose{\Animate(y)}
                         \For(y)(\Buy(x))(\We) :e\toQ e\toQ t$}
        \hspace{15em}}
    \qobitree
$}
\caption{\protect\phrase{What we bought~\_ for whom}, with the narrow-scope reading for
    \protect\phrase{whom}}
\label{f:what-we-bought-t-for-whom'}
\end{figure}

The first property is that \wh-phrases appear both raised and
in-situ.  For instance, the clauses in~\eqref{e:double-wh}
contain \phrase{what} raised and \phrase{whom} in-situ.
\ex. \label{e:double-wh}
     \a. \label{e:what-did-we-buy-t-for-whom}
         What did we buy~\_ for whom?
     \nopagebreak
     \b. \label{e:what-we-bought-t-for-whom}
         what we bought~\_ for whom\par
Ignoring the subject-auxiliary inversion
in~\eqref{e:what-did-we-buy-t-for-whom}, these two clauses are identical.
To derive them, all we need is an uncontroversial meaning
for \phrase{for}:
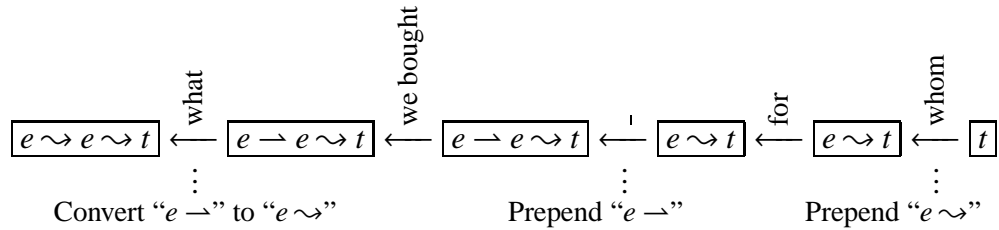
\begin{equation}
\label{e:for-denotation}
    \denote{\phrase{for}} = \For : e\toF(e\toF t)\toF(e\toF t).
\end{equation}
Given that it was with raised usage in mind that we assigned to
\phrase{whom} its meaning in~\eqref{e:who-denotation}, it may come as a
surprise that the same meaning works equally well for in-situ usage.  But
it does all work out: The derivation, which culminates in the top-level
type $e\toQ e\toQ t$, is shown in
Figure~\ref{f:what-we-bought-t-for-whom'}.  (If we
assume furthermore that \phrase{remember} can take semantic type $(e\toQ
e\toQ t)\toF e\toF t$, then it is straightforward to produce the
narrow-scope reading~\eqref{e:narrow-scope-answer-2} of the
example~\eqref{e:baker-indirect-2} from \sref{s:introduction}.)

To see how this derivation works, it is useful to
examine~\eqref{e:wh-denotations}, where \phrase{what} and \phrase{who} were
assigned the (polymorphic) type $(e\toC\gamma) \toF (e\toQ\gamma)$.
In~\sref{s:raised-wh-phrases}, this type was justified because
\phrase{what} needed to convert ($\toF$) an $e$\nobreakdash-\hspace{0pt}taking
continuation ($\toC$) to an $e$\nobreakdash-\hspace{0pt}wondering question ($\toQ$).
However, the same type can also be written as
$e\cc{\gamma}{e\toQ\gamma}$.
A value of this
type takes as input an $e$\nobreakdash-taking continuation whose answer is
$\gamma$, but returns a final answer of type $e\toQ\gamma$ instead.
Informally speaking, interrogative NPs
act like $e$s locally, but in addition prepends ``$e\>\toQ$''
to the current answer type.

Hence, as one might expect, the double-\wh\ constructions
in~\eqref{e:double-wh} receive the semantic type $e\toQ e\toQ t$.  The
first ``$e\>\toQ$'' in the type is contributed by the raised \wh-phrase
\phrase{what}, or rather, contributed as ``$e\>\toC$'' by the extraction
gap and subsequently converted to ``$e\>\toQ$'' by \phrase{what}.  The
second ``$e\>\toQ$'' in the type is contributed directly by the in-situ
\wh-phrase \phrase{whom}.  Figure~\ref{f:building-type} illustrates this
process.

\begin{figure}[tp]
\centerline{$
\newcommand{\explanation}[1]{\underset{\makebox[0pt]{\hss\small\strut#1\hss}}{\vdots}}
\newcommand{\stagetext}[2][]{\xleftarrow[#1]{\rotateleft{\mbox{\small\strut#2}}}}
    \boxed{e\toQ e\toQ t}
    \stagetext[\explanation{Convert ``$e\>\toC$'' to ``$e\>\toQ$''}]{what}
    \boxed{e\toC e\toQ t}
    \stagetext{we bought}
    \boxed{e\toC e\toQ t}
    \stagetext[\explanation{\hspace{-2em}Prepend ``$e\>\toC$''}]{\_}
    \boxed{e\toQ t}
    \stagetext{for}
    \boxed{e\toQ t}
    \stagetext[\explanation{\hspace{-3em}Prepend ``$e\>\toQ$''}]{whom}
    \boxed{t}
$}
\caption{Building the semantic type $e\toQ e\toQ t$ for
\protect\phrase{what we bought~\_ for whom}, with the narrow-scope reading for
\protect\phrase{whom}.  Whereas the in-situ elements~\protect\phrase{\_} and \protect\phrase{whom}
manipulate the answer type, the raised element \protect\phrase{what} manipulates
the value type.}
\label{f:building-type}
\end{figure}

It may be initially perplexing that the types in
Figure~\ref{f:building-type} are manipulated \emph{right-to-left}.  This
pattern is explained if we postulate that evaluation in natural
language tends to proceed \emph{left-to-right}.
That is, when a function occurs linearly before its argument, the
function-then-argument version of lifted function application is preferred,
and vice versa.
Left-to-right evaluation results in right-to-left answer type manipulation,
because constituents to the
left decide what the answer type looks like at outer levels.
For example, $\denote{\phrase{\_}}$ wants the answer type to be
a continuation at the outermost level it gets to affect.  When building the
answer type bottom-up from~$t$ to $e\toQ e\toQ t$, the outermost decisions
are executed last, not first.

Critical to the ability of the type $(e\toC\gamma) \toF (e\toQ\gamma)$ to
serve two roles at once is the lowering rule.  In-situ
\wh-phrases (and gaps) combine with other constituents and manipulate the
answer type through lifted function application.  By contrast, raised
\wh-phrases combine with other constituents through ordinary (unlifted) function
application, and perform the conversion from ``$e\>\toC$'' to ``$e\>\toQ$''
not on the answer type but on the value type.  Before a raised \wh-phrase
can act on a gapped clause, then, the clause needs a meaning whose value
type---\emph{not answer type}---is of the form $e\toC\dotsb$.  The lowering
rule fills this need: It extracts an answer out of a lifted value
by feeding it the identity continuation.

\subsection{Higher-Order Continuations}
\label{s:higher-order}

We have seen above that, in my analysis of interrogatives, a single
denotation for each \wh-phrase suffices for both raised and in-situ
appearances, as long as the \wh-phrase takes narrow scope.  To fulfill the
promises I made in \sref{s:introduction}, I have to turn to wide scope and
account for two additional facts about interrogatives: First, I have to
show in my system that in-situ \wh-phrases can take semantic scope wider
than the immediately enclosing clause, as they do in my initial examples
\eqref{e:wide-scope-answer-1} and~\eqref{e:wide-scope-answer-2}.  Second,
I have to show that raised \wh-phrases cannot take wide scope; in other
words, they must take semantic scope exactly where they are overtly
located.

I claim that \emp{higher-order continuations} account for wide scope
interrogatives\hspace{0pt}.%
  \footnote{In the same spirit, \citet{barker-higher-order} used
  higher-order continuations to treat wide-scope specific indefinites and
  interactions between coordination and antecedent-contained deletion.}
Our analyses in previous sections are lifted only to the first order.  In
PTQ terms, this means that our values are sets of sets; more generally, our
types are of the form~$\alpha\cc{\gamma}{\gamma'}$.  Wide scope calls for
lifting to the second order.  In PTQ terms, this means that our values need
to be sets of sets of sets of sets; more generally, we need to deal with
types of the form~$\alpha\cc{\gamma}{\gamma'}\cc{\delta}{\delta'}$.  Recall
that values lifted to the first order are manipulated using four additional
semantic rules: lifting, lowering, and lifted function application (two
versions).  I explain below how to introduce further semantic rules into
the grammar that manipulate values lifted to higher orders.

As described in \sref{s:generalization-1}, lifted function application is
obtained by ``lifting'' ordinary function application.  Since ordinary
function application is a binary rule, it can be lifted in two ways
(evaluation orders), giving two rules for lifted function application.  In
general, any $n$\nobreakdash-ary semantic rule, schematically
\begin{equation}
\label{e:unlifted-rule}
    \leaf{$x_1 : \alpha_1$}
    \leaf{$x_2 : \alpha_2$}
    \leaf{$\dotsm$}
    \leaf{$x_n : \alpha_n$}
    \branch{4}{$y : \beta$}
    \qobitreecenter[,]
\end{equation}
can be lifted in $n!$ ways, giving rise to $n!$ lifted rules: For each
permutation~$\sigma$ of the numbers $1$, $2$,~\dots,~$n$, 
we can lift~\eqref{e:unlifted-rule} to a new rule
\begin{equation}
\label{e:lifted-rule}
    \leaf{$\bar{x}_1 : \alpha_1\cc{\gamma_{\sigma_1}}{\gamma_{\sigma_1-1}}$}
    \leaf{$\bar{x}_2 : \alpha_2\cc{\gamma_{\sigma_2}}{\gamma_{\sigma_2-1}}$}
    \leaf{$\dotsm$}                                                    
    \leaf{$\bar{x}_n : \alpha_n\cc{\gamma_{\sigma_n}}{\gamma_{\sigma_n-1}}$}
    \branch{4}{$
        \fun{c}
        \bar{x}_{\sigma^{-1}_1}\bigl(\fun{x_{\sigma^{-1}_1}}
        \bar{x}_{\sigma^{-1}_2}\bigl(\fun{x_{\sigma^{-1}_2}}
        \dotsm
        \bar{x}_{\sigma^{-1}_n}\bigl(\fun{x_{\sigma^{-1}_n}}
        c(y)
        \bigr)
        \dotsm
        \bigr)
        \bigr)
        : \beta\cc{\gamma_n}{\gamma_0}$}
    \qobitreecenter[.]
\end{equation}
For example, ordinary function application
(Figure~\ref{f:direct-application}) can be lifted with $\sigma_1=1$,
$\sigma_2=2$ (Figure~\ref{f:lifted-fa-1}) or with $\sigma_1=2$,
$\sigma_2=1$ (Figure~\ref{f:lifted-fa-2}).

Let $G$ be a Montague grammar, each of whose semantic rules are of the form
in~\eqref{e:unlifted-rule}.  We can \emp{lift}~$G$ to a new
grammar~$G'$, with the following semantic rules:
\begin{itemize}
\item the value lifting rule (Figure~\ref{f:lifting});
\item the value lowering rule (Figure~\ref{f:lowering}); and
\item every rule in~$G$, along with the $n!$ ways to lift it, where $n$ is
      the arity.
\end{itemize}
Let $G_0$ be ``pure'' Montague grammar, where the only semantic rule
is function application.  Lifting
$G_0$ gives a new grammar $G_0'$; call it~$G_1$.  Lifting $G_1$ gives
another grammar $G_1' = G_0''$; call it~$G_2$.
\begin{table}
\caption{Lifting Montague grammar.  Starting from function application
($G_0$), lifting once gives a grammar ($G_1$) with $2$ unary rules and $3$
binary rules.  Lifting again gives a grammar ($G_2$) with $4$ unary rules
and $7$ binary rules.  In the table, ``$f>x$'' means ``evaluating function
first''; ``$x>f$'' means ``evaluating argument first''.}
\label{f:lifting-grammar}
\def\=#1 {\node{#1}{$\bullet$}}
\def\fta{, lifted $f>x$}
\def\atf{, lifted $x>f$}
\begin{center}
\begin{tabular}{lccccc}
\hline
Unary rule & $G_0$ && $G_1$ && $G_2$ \\
\hline
Value lifting                         &           & & \=lift1   &=& \=lift2   \\
Value lifting, lifted                 &           & &           & & \=liftL2   \\
Value lowering                        &           & & \=lower1  &=& \=lower2   \\
Value lowering, lifted                &           & &           & & \=lowerL2   \\
\hline
Binary rule & $G_0$ && $G_1$ && $G_2$ \\
\hline
Function application\fta\fta          &           & &           & & \=faFF2   \\
Function application\fta              &           & & \=faF1    &=& \=faF2   \\
Function application\fta\atf          &           & &           & & \=faFA2   \\
Function application                  & \=fa0     &=& \=fa1     &=& \=fa2   \\
Function application\atf\fta          &           & &           & & \=faAF2   \\
Function application\atf              &           & & \=faA1    &=& \=faA2   \\
Function application\atf\atf          &           & &           & & \=faAA2   \\
\hline
\end{tabular}
\anodeconnect[r]{lift1}[l]{liftL2}%
\anodeconnect[r]{lower1}[l]{lowerL2}%
\anodeconnect[r]{fa0}[l]{faF1}%
\anodeconnect[r]{fa0}[l]{faA1}%
\anodeconnect[r]{faF1}[l]{faFF2}%
\anodeconnect[r]{faF1}[l]{faFA2}%
\anodeconnect[r]{faA1}[l]{faAF2}%
\anodeconnect[r]{faA1}[l]{faAA2}%
\end{center}
\end{table}
These grammars and their rules are illustrated in
Figure~\ref{f:lifting-grammar}.  The grammar~$G_2$ contains the semantic
rules we need to manipulate values lifted to the second order: lifted
lifting, lifted lowering, and twice-lifted function application (four
versions).  The process may continue indefinitely.

In a once-lifted grammar, many types are of the
form~$\alpha\cc{\gamma}{\gamma'}$.
As explained in \sref{s:extraction}, such a type can be understood
to mean ``acts locally like an~$\alpha$ while changing the
answer type from~$\gamma$ to~$\gamma'$''.  In a twice-lifted grammar, many
types are of the form~$\alpha\cc{\gamma}{\gamma'}\cc{\delta}{\delta'}$.
One way to understand such types is to think of a derivation in
a twice-lifted grammar as maintaining two answer types---an \emph{inner}
answer type corresponding to the first time the grammar is lifted, and an
\emph{outer} one corresponding to the second time.  A type of the
form~$\alpha\cc{\gamma}{\gamma'}\cc{\delta}{\delta'}$ means ``acts locally
like an~$\alpha$ while changing the inner answer type
from~$\gamma$ to~$\gamma'$ and the outer answer type from~$\delta$
to~$\delta'$''.

To strengthen this understanding, let us examine how answer types are
manipulated in the four binary composition rules that result from\pagebreak[3]
lifting function application twice.  In Figure~\ref{f:lifted-lifted-fa},
I expand out the rules, in particular the types.
\begin{figure}[tp]
\centerline{$
    \leaf{$\bar{\bar{f}} : (\alpha\toF\beta)
               \cc{\gamma_1}{\gamma_0}
               \cc{\delta_1}{\delta_0}$}
    \leaf{$\bar{\bar{x}} : \alpha
               \cc{\gamma_2}{\gamma_1}
               \cc{\delta_2}{\delta_1}$}
    \binarybranch[>^>]{\fun{d}
                  \bar{\bar{f}}\bigl(\fun{\bar{f}}
                  \bar{\bar{x}}\bigl(\fun{\bar{x}}
                  d\bigl(\fun{c}
                  \bar{f}\bigl(\fun{f}
                  \bar{x}\bigl(\fun{x}
                  c(f(x))\bigr)\bigr)\bigr)\bigr)\bigr)
        : \beta\cc{\gamma_2}{\gamma_0}
               \cc{\delta_2}{\delta_0}}
    \qobitreecenter
$}
\vspace{\jot}
\centerline{$
    \leaf{$\bar{\bar{f}} : (\alpha\toF\beta)
               \cc{\gamma_1}{\gamma_0}
               \cc{\delta_2}{\delta_1}$}
    \leaf{$\bar{\bar{x}} : \alpha
               \cc{\gamma_2}{\gamma_1}
               \cc{\delta_1}{\delta_0}$}
    \binarybranch[>^<]{\fun{d}
                  \bar{\bar{x}}\bigl(\fun{\bar{x}}
                  \bar{\bar{f}}\bigl(\fun{\bar{f}}
                  d\bigl(\fun{c}
                  \bar{f}\bigl(\fun{f}
                  \bar{x}\bigl(\fun{x}
                  c(f(x))\bigr)\bigr)\bigr)\bigr)\bigr)
        : \beta\cc{\gamma_2}{\gamma_0}
               \cc{\delta_2}{\delta_0}}
    \qobitreecenter
$}
\vspace{\jot}
\centerline{$
    \leaf{$\bar{\bar{f}} : (\alpha\toF\beta)
               \cc{\gamma_2}{\gamma_1}
               \cc{\delta_1}{\delta_0}$}
    \leaf{$\bar{\bar{x}} : \alpha
               \cc{\gamma_1}{\gamma_0}
               \cc{\delta_2}{\delta_1}$}
    \binarybranch[<^>]{\fun{d}
                  \bar{\bar{f}}\bigl(\fun{\bar{f}}
                  \bar{\bar{x}}\bigl(\fun{\bar{x}}
                  d\bigl(\fun{c}
                  \bar{x}\bigl(\fun{x}
                  \bar{f}\bigl(\fun{f}
                  c(f(x))\bigr)\bigr)\bigr)\bigr)\bigr)
        : \beta\cc{\gamma_2}{\gamma_0}
               \cc{\delta_2}{\delta_0}}
    \qobitreecenter
$}
\vspace{\jot}
\centerline{$
    \leaf{$\bar{\bar{f}} : (\alpha\toF\beta)
               \cc{\gamma_2}{\gamma_1}
               \cc{\delta_2}{\delta_1}$}
    \leaf{$\bar{\bar{x}} : \alpha
               \cc{\gamma_1}{\gamma_0}
               \cc{\delta_1}{\delta_0}$}
    \binarybranch[<^<]{\fun{d}
                  \bar{\bar{x}}\bigl(\fun{\bar{x}}
                  \bar{\bar{f}}\bigl(\fun{\bar{f}}
                  d\bigl(\fun{c}
                  \bar{x}\bigl(\fun{x}
                  \bar{f}\bigl(\fun{f}
                  c(f(x))\bigr)\bigr)\bigr)\bigr)\bigr)
        : \beta\cc{\gamma_2}{\gamma_0}
               \cc{\delta_2}{\delta_0}}
    \qobitreecenter
$}
\caption{Twice-lifted function application (four versions)}
\label{f:lifted-lifted-fa}
\end{figure}
%
In the rules $>^>$ and~$>^<$, evaluation proceeds from function to argument\pagebreak[3]
at the first continuation level.  Accordingly, the subscripts on~$\gamma$
show that the inner answer type is threaded first through the
argument~$\smash[t]{\bar{\bar{x}}}$ and then through the function~$\smash[t]{\bar{\bar{f}}}$.  In
the rules $<^>$ and~$<^<$ the reverse happens: Evaluation proceeds from
argument to function, and the inner answer type is threaded through
first~$\smash[t]{\bar{\bar{f}}}$ and then~$\smash[t]{\bar{\bar{x}}}$.

Similarly for the second continuation level: In the rules $>^>$ and~$<^>$,
evaluation proceeds from function to argument, and the subscripts
on~$\delta$ show that the outer answer type is threaded first
through~$\smash[t]{\bar{\bar{x}}}$ and then through~$\smash[t]{\bar{\bar{f}}}$.  In the rules
$>^<$ and~$<^<$, evaluation proceeds from argument to function, and the
outer answer type is threaded through first~$\smash[t]{\bar{\bar{f}}}$ and
then~$\smash[t]{\bar{\bar{x}}}$.

\subsection{Wide Scope and Baker's Ambiguity}
\label{s:wide-scope}

With second-order continuations, we can compute the wide-scope
reading \eqref{e:wide-scope-answer-2} for
the sentence~\eqref{e:baker-indirect-2} from~\sref{s:introduction},
repeated here with gaps represented explicitly.
\ex.[(\pref{e:baker-indirect-2})]
    \label{e:baker-indirect-2'}
    Who do you think~\_ remembers [what we bought~\_ for whom]?\par
I extend the grammar only by lifting it again as discussed above, and
add no new denotations to the lexicon other than the obvious missing
entries
\begin{equation}
    \You:e,\qquad \Think:t\toF e\toF t.
\end{equation}

The wide-scope reading of the example~\eqref{e:baker-indirect-2} is a
double-\wh\ question.  Thus, we expect the matrix clause to have the semantic
type $e\toQ e\toQ t$.  What about the embedded clause?  It acts locally
like a single-\wh\ question (\phrase{what}), but in addition should prepend
``$e\>\toQ$'' to the answer type (\phrase{whom}).  We thus expect the
embedded clause to have the type \smash[b]{$(e\toQ
t)\cc{\delta}{e\toQ\delta}$}, a special case of which is $(e\toQ
t)\cc{t}{e\toQ t}$.

\begin{figure}
\centerline{$
    \leaf{$
      \begin{stacked}
        \displayleft{\denote{\phrase{what}}: (e\toC t)}
        \displayright{\toF(e\toQ t)}
      \end{stacked}$}
    \unarybranch[\wedge]{\bigl((e\toC t) \toF(e\toQ t)\bigr) \cc{e\toQ \delta}{e\toQ \delta}}
    \leaf{$\We : e$}
    \unarybranch[\wedge]{e\cc{e\toC t}{e\toC t}}
    \unarybranch[\wedge]{e\cc{e\toC t}{e\toC t}\cc{e\toQ \delta}{e\toQ \delta}}
    \leaf{$\Buy : e\toF e\toF t$\hspace*{-0.05in}}
    \unarybranch[\wedge]{(e\toF e\toF t)\cc{e\toC t}{e\toC t}}
    \unarybranch[\wedge]{(e\toF e\toF t) $\\$ \cc{e\toC t}{e\toC t}\cc{e\toQ \delta}{e\toQ \delta}}
    \leaf{$\fun{c}c : e\cc{t}{e\toC t}$}
    \unarybranch[\wedge]{e\cc{t}{e\toC t}\cc{e\toQ \delta}{e\toQ \delta}}
    \binarybranch[>^>]{\fun{d} d\bigl(\fun{c} \fun{x} c(\Buy(x))\bigr)
        $\\$\qquad :(e\toF t)\cc{t}{e\toC t}\cc{e\toQ \delta}{e\toQ\delta}}
    \leaf{$\For : e\toF (e\toF t) \toF (e\toF t)$}
    \unarybranch[\wedge]{\bigl(e\toF (e\toF t) \quad $\\$ \toF (e\toF t)\bigr)\cc{t}{t}}
    \unarybranchfakewidth[\wedge]{\bigl(e\toF (e\toF t) \quad\qquad $\\$ \toF (e\toF t)\bigr)\cc{t}{t}\cc{e\toQ \delta}{e\toQ \delta}}
    \leaf{$\denote{\phrase{who}}\quad $\\$ : e\cc{\delta}{e\toQ \delta}$}
    \unarybranch[\wedge^*]{e\cc{t}{t}\cc{\delta}{e\toQ \delta}}
    \binarybranchfakewidth[>^>]{\qquad\bigl((e\toF t)\toF(e\toF t)\bigr)\cc{t}{t}\cc{\delta}{e\toQ \delta}\qquad}
    \binarybranchfakewidth[>^>]{
        \rlap{$\fun{d} \fun{y} \presuppose{\Animate(y)} d\bigl(\fun{c} \fun{x} $}\hspace{11em}$\\$
        \rlap{$c(\For(y)(\Buy(x)))\bigr) : (e\toF t)\cc{t}{e\toC t}\cc{\delta}{e\toQ \delta}$}\hspace{7em}}
    \binarybranchfakewidth[>^>]{
        \rlap{$\fun{d} \fun{y} \presuppose{\Animate(y)} d\bigl(\fun{c} \fun{x} $}\qquad\qquad$\\$
        \rlap{$c(\For(y)(\Buy(x))(\We))\bigr) : t\cc{t}{e\toC t}\cc{\delta}{e\toQ \delta}$}\qquad}
    \unarybranchfakewidth[\vee^*]{
        \rlap{$\fun{d} \fun{y} \presuppose{\Animate(y)} d\bigl(\fun{x} $}\qquad\qquad$\\$
        \rlap{$\For(y)(\Buy(x))(\We)\bigr) : (e\toC t)\cc{\delta}{e\toQ \delta}$}\qquad}
    \binarybranch[>]{
      \rlap{$\fun{d} \fun{y} \presuppose{\Animate(y)} d\bigl(\fun{x} \presuppose{\neg\Animate(x)} $}\qquad\qquad$\\$
      \rlap{$\For(y)(\Buy(x))(\We)\bigr) : (e\toQ t)\cc{\delta}{e\toQ \delta}$}\qquad}
    \qobitree
    \hspace{-1.75em}
$}
\caption{\protect\phrase{What we bought~\_ for whom}, with the wide-scope
    reading for \protect\phrase{whom}}
\label{f:what-we-bought-t-for-whom-wide}
\end{figure}

Figure~\ref{f:what-we-bought-t-for-whom-wide} shows a derivation for
\phrase{what we bought~\_ for whom} that culminates in precisely the
expected type.
The interesting part of the derivation is how the three
continuation-manipulating elements \phrase{what},~\phrase{\_}, and
\phrase{whom} enter it.  The narrow-scope elements \phrase{what}
and~\phrase{\_} need to manipulate the inner answer type while remaining
oblivious to the second continuation level, so we lift them.  The
wide-scope element \phrase{whom}, on the other hand, needs to manipulate
the outer answer type while leaving the first continuation
level alone; to achieve this effect, we lift it ``from the inside'' using the
lifted lifting rule (depicted as $\wedge^*$).  Near the top of the
derivation, we use the lifted lowering rule (depicted as $\vee^*$)
to lower the embedded clause's denotation ``from the inside'', that is, on the
first rather than second continuation level.

Given a meaning for \phrase{what we bought~\_ for whom} of the expected
type above, the semantics of the matrix question follows easily from the
techniques already demonstrated in previous sections.  Second-order
continuations are no longer involved.  The embedded clause is just
a constituent that contains an in-situ \wh-phrase \phrase{whom}; like any
other such constituent, it combines with the rest of the sentence,
including the raised \wh-phrase \phrase{who}, to give a final denotation of
type $e\toQ e\toQ t$, typical of a double-\wh\ interrogative.
Figure~\ref{f:who-you-think-t-remembers-what-we-bought-t-for-whom-wide}
shows the derivation.

\begin{figure}
\centerline{$
    \leaf{$
      \begin{stacked}
        \displayleft{\denote{\phrase{who}}}
        \displayright{: (e\toC e\toQ t)}
        \displayright{\toF(e\toQ e\toQ t)}
      \end{stacked}\hspace{-1em}$}
    \leaf{$\You:e$}
    \unarybranchfakewidth[\wedge]{e\cc{e\toC e\toQ t}{e\toC e\toQ t}}
    \leaf{\qquad\llap{$\Think:t\toF e\toC t$}}
    \unarybranchfakewidth[\wedge]{\qquad\llap{$(t\toF e\toC t)\cc{e\toC e\toQ t}{e\toC e\toQ t}$}}
    \leaf{\qquad\llap{$\fun{c}c:e\cc{e\toQ t}{e\toC e\toQ t}$}}
    \leaf{\qquad\llap{$\Remember:(e\toQ t)\toF e\toF t$}}
    \unarybranchfakewidth[\wedge]{\qquad\llap{$\bigl((e\toQ t)\toF e\toF t\bigr)\cc{e\toQ t}{e\toQ t}$}}
    \roof{what we bought~\_ for whom\\(Figure~\ref{f:what-we-bought-t-for-whom-wide})}{$
        (e\toQ t)\cc{t}{e\toQ t}
      $}
    \binarybranchfakewidth[>]{(e\toF t)\cc{t}{e\toQ t}}
    \binarybranchfakewidth[>]{
      \rlap{$\fun{d} \fun{z} \fun{y} \presuppose{\Animate(y)}$}\qquad\qquad\qquad$\\$
      \rlap{$d\bigl( \Remember\bigl( \fun{x} \presuppose{\neg\Animate(x)} $}\qquad\qquad$\\$
      \rlap{$\For(y)(\Buy(x))(\We)\bigr)(z) \bigr) : t\cc{t}{e\toC e\toQ t}$}\qquad}
    \binarybranchfakewidth[>]{
      \rlap{$\fun{d} \fun{z} \fun{y} \presuppose{\Animate(y)}$}\qquad\qquad\qquad$\\$
      \rlap{$d\bigl( \Think\bigl( \Remember\bigl( \fun{x} \presuppose{\neg\Animate(x)} $}\qquad\qquad$\\$
      \rlap{$\For(y)(\Buy(x))(\We)\bigr)(z) \bigr) \bigr) : (e\toF t)\cc{t}{e\toC e\toQ t}$}\qquad}
    \binarybranchfakewidth[>]{
      \rlap{$\fun{d} \fun{z} \fun{y} \presuppose{\Animate(y)}$}\qquad\qquad\qquad$\\$
      \rlap{$d\bigl( \Think\bigl( \Remember\bigl( \fun{x} \presuppose{\neg\Animate(x)} $}\qquad\qquad$\\$
      \rlap{$\For(y)(\Buy(x))(\We)\bigr)(z) \bigr)(\You) \bigr) : t\cc{t}{e\toC e\toQ t}$}\qquad}
    \unarybranchfakewidth[\vee]{
      \rlap{$\fun{z} \fun{y} \presuppose{\Animate(y)}$}\qquad\qquad\qquad$\\$
      \rlap{$\Think\bigl( \Remember\bigl( \fun{x} \presuppose{\neg\Animate(x)} $}\qquad\qquad$\\$
      \rlap{$\For(y)(\Buy(x))(\We)\bigr)(z) \bigr)(\You) : e\toC e\toQ t$}\qquad}
    \binarybranchfakewidth{
      \rlap{$\fun{z} \presuppose{\Animate(z)} \fun{y} \presuppose{\Animate(y)}$}\qquad\qquad\qquad$\\$
      \rlap{$\Think\bigl( \Remember\bigl( \fun{x} \presuppose{\neg\Animate(x)} $}\qquad\qquad$\\$
      \rlap{$\For(y)(\Buy(x))(\We)\bigr)(z) \bigr)(\You) : e\toQ e\toQ t$}\qquad}
    \qobitree
    \hspace{3em}
$}
\caption{\protect\phrase{Who you thinks~\_ remembers what we bought~\_ for whom}, with the wide-scope
    reading for \protect\phrase{whom}}
\label{f:who-you-think-t-remembers-what-we-bought-t-for-whom-wide}
\end{figure}

We have seen that higher-order continuations allow in-situ
\wh-phrases to take wide scope.  I now explain why raised
\wh-phrases cannot take wide scope,
no matter how many times we lift the grammar.  What does it mean for
a raised \wh-phrase to take wide scope?  Based on the
analyses so far, I make the following definitional characterization: A
raised \wh-phrase takes narrow scope when it contributes its ``$e\>\toQ$''
to the clause's value type, and wide scope when it contributes its
``$e\>\toQ$'' to the clause's outgoing answer type (or rather, one of the
clause's outgoing answer types, in the presence of higher-order
continuations).
More precisely, consider a clause with a raised
\wh-phrase in front and a corresponding gap inside.
\exi. \label{e:wh-clause}
      [$a$ [$b$ wh] [$c$ {\dots} [$d$ t] \mbox{ \dots}]]\par
For the \wh-phrase $b$ to take narrow scope is
for the clause $a$ to have semantic type of the form
$   \bigl(e\toQ\alpha\bigr)
    \smash{\cc{\gamma_1}{\gamma'_1} \dotsm \cc{\gamma_n}{\gamma'_n}} $,
and the clause-sans-\wh-phrase $c$ the corresponding form
$   \bigl(e\toC\alpha\bigr)
    \smash{\cc{\gamma_1}{\gamma'_1} \dotsm \cc{\gamma_n}{\gamma'_n}} $,
such that the ``$e\>\toQ$'' was contributed as ``$e\>\toC$'' by the gap $d$
and subsequently converted to ``$e\>\toQ$'' by the \wh-phrase $b$.  This is
demonstrated in \sref{s:raised-wh-phrases}.

By contrast, for the \wh-phrase $b$ to take wide scope is for $a$
to have semantic type of the form
$   \beta
    \cc{\alpha}{e\toQ\alpha}
    \smash{\cc{\gamma_1}{\gamma'_1} \dotsm \cc{\gamma_n}{\gamma'_n}} $,
and $c$ the corresponding form
$   \beta
    \cc{\alpha}{e\toC\alpha}
    \smash{\cc{\gamma_1}{\gamma'_1} \dotsm \cc{\gamma_n}{\gamma'_n}} $,
such that the ``$e\>\toQ$'' was contributed as ``$e\>\toC$'' by $d$ and
subsequently converted to ``$e\>\toQ$'' by $b$.  This
is impossible because it requires replacing ``$e\>\toC$'' with
``$e\>\toQ$'' in an answer type, a feat performed by neither any element in
the lexicon nor any rule in the grammar:  A survey of
the lexical items and grammar rules in this paper reveals that the they and
their descendants-by-lifting only manipulate answer types by adding
to them.  Nothing ever takes apart any answer type
until the answer has been lowered to value level.  The
formalism I use here does not stipulate this---in fact, one can easily
introduce into the lexicon raised \wh-phrases that take wide scope, but
English does not appear to contain these denotations:
\begin{subequations}
\begin{alignat}{2}
\denote{\phrase{what}}
       &\ne \fun{p} \fun{c} \fun{x} \presuppose{\neg\Animate(x)} p(c)(x)
      &&\eqishcolon\textstyle
          \alpha\cc{\gamma}{e\toC\gamma} \toF
          \alpha\cc{\gamma}{e\toQ\gamma}, \\
\denote{\phrase{who}}
       &\ne \fun{p} \fun{c} \fun{x} \presuppose{    \Animate(x)} p(c)(x)
      &&\eqishcolon\textstyle
          \alpha\cc{\gamma}{e\toC\gamma} \toF
          \alpha\cc{\gamma}{e\toQ\gamma}.
\end{alignat}
\end{subequations}

\section{Discussion}
\label{s:discussion}

This paper presents a grammar fragment that captures two properties
of English interrogatives:  First, \wh-phrases can appear both raised and
in-situ.  Second, in-situ \wh-phrases can take scope beyond the immediately
enclosing clause, but raised \wh-phrases must take scope exactly where they
are pronounced.  My fragment is inspired by Barker's continuation
semantics for natural
language~\citeyearpar{barker-continuations,barker-higher-order} and
work on continuations and typed contexts in programming languages
\citep{danvy-functional,danvy-abstracting,murthy-control,wadler-composable}.

As a semantics of interrogatives, this paper leaves many concerns
unaddressed:
languages other than English;
pair-list questions;
relative clauses;
interactions with intensionality and quantification; and so on.
Regardless, the two properties of interrogatives that I
do explore fall out surprisingly naturally:
The same \wh-denotation works both raised and in-situ,
but in raised position it is forced
to take overt scope by a theorem of the type system.
The basic ideas probably carry over to other kinds of
so-called \=A-movement, such as topicalization.  Continuations also suggest
new ways to understand phenomena such as superiority and pied-piping, but
for lack of space I leave these investigations for elsewhere.

\theendnotes

\bibliographystyle{mcbride}
\bibliography{ccshan}

\end{document}